\documentclass[letterpaper,10pt,conference]{ieeeconf}
\IEEEoverridecommandlockouts
\usepackage{booktabs}
\usepackage{times}
\usepackage{amsmath,amssymb,mathtools}
\usepackage{graphicx}
\usepackage{xcolor}
\usepackage{microtype}
\usepackage{hyperref}
\usepackage{cite}
\usepackage{float}
\usepackage[ruled,vlined,linesnumbered]{algorithm2e}
\SetAlgoCaptionSeparator{.}
\SetKwInput{KwInput}{Input}
\SetKwInput{KwOutput}{Output}

\SetKwComment{Comment}{$\triangleright$\ }{}
\SetAlFnt{\footnotesize}
\DontPrintSemicolon

\newcommand{\sigmoid}[3]{\frac{1}{1+\exp\!\left(-#1\left(#2-#3\right)\right)}}

\graphicspath{{figs/}{./}}

\title{\LARGE \bf
A Robust Antenna Provides Tactile Feedback in a Multi-legged Robot
}
\author{Zhaochen J. Xu$^{1}$, Juntao He$^{1}$, Delfin Aydan$^{1}$, Malaika Taylor$^{1}$,%
Tianyu Wang$^{1}$, Jianfeng Lin$^{1}$,\\ Wesley Dyer$^{1}$, Daniel I. Goldman$^{1}$%
\thanks{$^{1}$All authors are with the Georgia Institute of Technology, Atlanta, GA 30332, USA.
{\tt\small \{zxu699, jhe391, daydan3, mtaylor389, tianyuwang, jianf.lin, wdyar6\}@gatech.edu, daniel.goldman@physics.gatech.edu}}%
}

\begin{document}
\maketitle
\begin{abstract}
Multi-legged elongate robots hold promise for maneuvering through complex environments. Prior work has demonstrated that reliable locomotion can be achieved using open-loop body undulation and foot placement on rugose terrain. However, robust navigation through confined spaces remains challenging when body-environment contact is extensive and terrain rheology varies rapidly. To address this challenge, we develop a pair of tactile antennae for multi-legged robots that enable real-time sensing of surrounding geometry, modeling the morphology and function of biological centipede antennae. Each antenna features gradient compliance, with a stiff base and soft tip, allowing repeated deformation and elastic recovery. Robophysical experiments reveal a relationship between continuous antenna curvature and contact force, leading to a simplified mapping from antenna deformation to inferred discrete collision states. We incorporate this mapping into a controller that selects among a set of locomotor maneuvers based on the inferred collision state. Experiments in obstacle-rich and confined environments demonstrate that tactile feedback enables reliable steering and allows the robot to recover from near-stuck conditions without requiring global environmental information or real-time vision. These results highlight how mechanically tuned tactile appendages can simplify sensing and enhance autonomy in elongate multi-legged robots operating in constrained spaces.

\end{abstract}

\begin{figure}[!htbp]
    \centering
    \includegraphics[width=0.98 \linewidth]{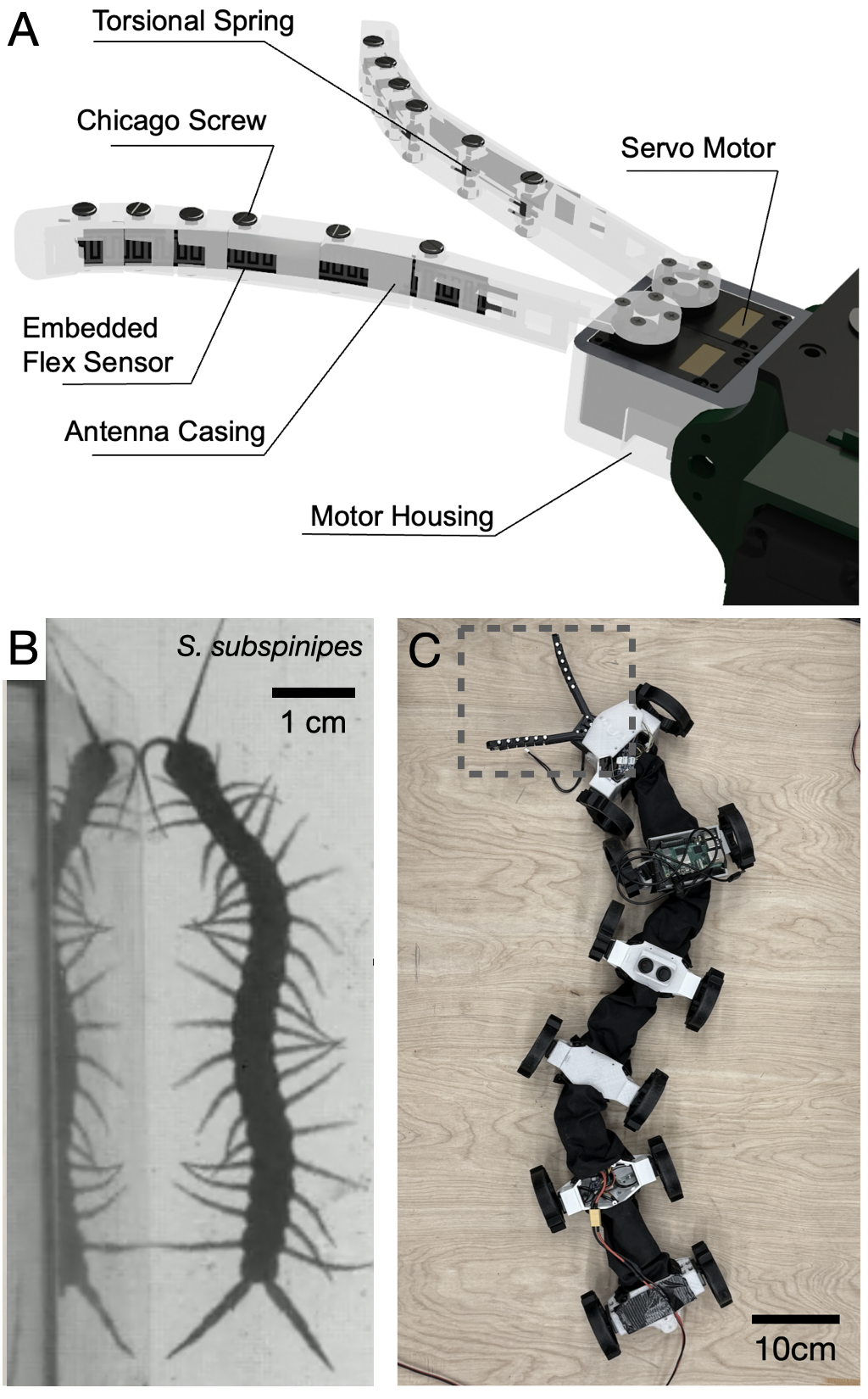}
    \caption{
    \textbf{Robophysical antenna model integrated on a multi-legged robot (top-down view).}
    (A) Antenna design; The antenna enables compliant contact sensing through measured bending.
    (B) Top-down view of the centipede \textit{S. subspinipes} showing wall-sensing of antenna.
    (C) Top-down view of the multi-legged robot SCUTTLE (Ground Control Robotics, Inc.). The dashed box highlights the modular antenna mounted on the anterior segment.
   }
    \label{fig:robot_overview}
\end{figure}

\section{Introduction}

Elongate multi-legged robots, characterized by segmented bodies and repeated simple legs\cite{chong2023multilegged,ozkan2021self,chong2023self}, provide an alternative route to robust terrestrial locomotion on complex terrain. Compared to common solutions such as bipedal and quadrupedal robots, their morphology enables redundancy and distributed contact interactions, while coordinated limb stepping and traveling waves of body undulation allow reliable mobility across heterogeneous substrates\cite{chong2023multilegged,he2025probabilistic, chong2023self,chong2022general}. As a result, stable locomotion can emerge with minimal sensing and limited computational control.

However, for field deployment, reliable mobility requires more than consistent forward locomotion: elongate multi-legged robots must regulate heading and make steering decisions while negotiating contact with surrounding terrain and obstacles\cite{chong2025omega,aoi2022advanced}. Small interaction-induced perturbations can accumulate, leading to deviations in heading or loss of forward progress under open-loop control. This challenge highlights the need for sensing strategies that remain effective during sustained contact and in visually occluded or cluttered environments. In such contact-rich settings, low-dimensional contact cues provide more direct guidance for control than reconstructing high-dimensional state, motivating simple tactile sensing as a practical and robust solution.

Biological systems suggest similar strategies based on contact sensing. Many arthropods use bilateral antennae to probe obstacles and guide steering through touch. Classic studies of cockroach wall-following demonstrated that stable, rapid contact-guided steering can be achieved using task-level feedback derived from antennal signals \cite{camhi1999high,cowan2006task,lee2008templates}. Subsequent work connected antenna mechanics, sensing, and body dynamics through the templates-and-anchors framework \cite{lee2008templates,full1999templates}, showing how antennal mechanics and reconfiguration shape sensory signals and simplify control during rapid navigation \cite{mongeau2014mechanical,mongeau2013locomotion,mongeau2015sensory}. Additional insect studies highlight how antennal structure and active touch behaviors influence tactile orientation and localization \cite{okada2000role,solomon2006robotic,loudon2014cricket}. These findings demonstrate that antenna contact can provide low-dimensional cues sufficient to guide locomotion decisions.
\begin{figure}[!tbp]
    \centering
    \includegraphics[width=0.95\linewidth]{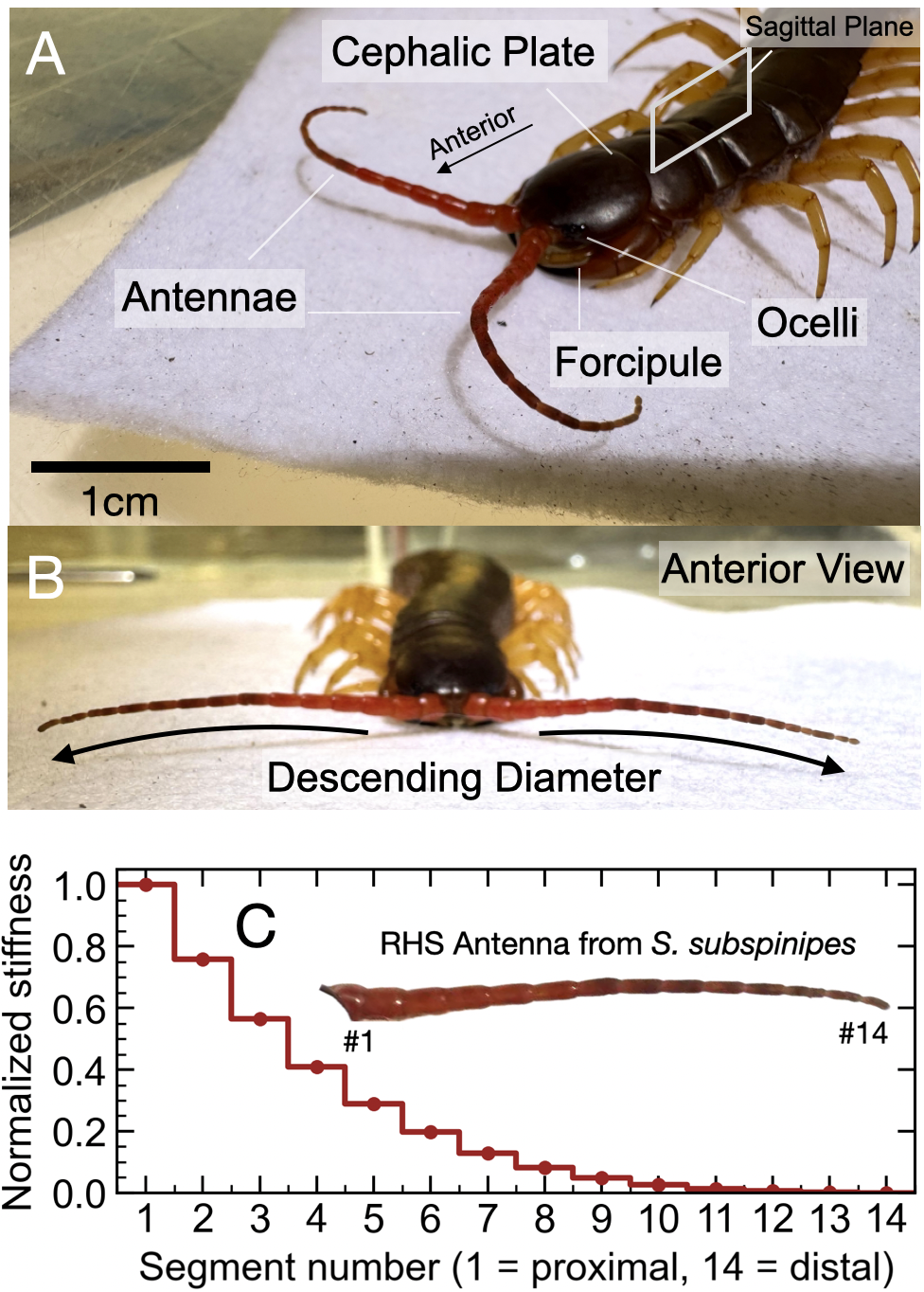}
    \caption{\textbf{Antenna morphology of a centipede.}
(A) Lateral view showing the cephalic plate, antennae, ocelli, and forcipules, with anterior direction and sagittal plane indicated (scale bar: 1 cm). (B) Anterior view highlighting the bilaterally extended antennae and their descending segment diameter from base to tip (proximal diameter 0.15 cm; distal diameter 0.02 cm). (C) Diameter based stiffness proxy along the antenna, computed from the proximal to distal diameter profile using a fourth power scaling ($\hat{K}_i \propto d^{4}$), showing a descending stiffness distribution from base to tip.}
    \label{fig:niceshot}
\end{figure}
Roboticists have explored similar ideas by developing tactile antennae and whisker-like sensors that convert contact state into feedback for locomotion. Early work demonstrated contact-guided wall-following using passive tactile sensing and dynamical control \cite{lamperski2005dynamical}. To isolate how mechanics shapes what is sensed, tunable physical models and multi-segment robotic antennae were developed to vary stiffness and segmentation under controlled contact \cite{demir2010tunable,mongeau2014mechanical}. In addition, whiskered robotic platforms\cite{pearson2007whiskerbot,yu2024whisker,solomon2006robotic} showed that active tactile sensing can provide navigation-relevant cues, such as radial distance estimation, while base signals can detect local features during contact \cite{evans2013effect,solomon2006robotic,pearson2011biomimetic}. More recent systems expand the tactile sensing toolbox, including vision-based whisker arrays, triboelectric bionic antenna sensors, and compact multi-bend shape sensing methods for slender structures \cite{kent2021whisksight,zhu2023self,shahmiri2020sharc,yu2022bioinspired}. Insect-scale robophysical antenna platforms such as CITRAS further demonstrate how segmented structures and distributed sensing support distance and gap-width estimation through touch \cite{mcdonnell2025design}. Tactile sensing has also been shown to enhance locomotion capabilities in elongate robots \cite{he2025probabilistic}, enabling behaviors such as vertical obstacle negotiation through contact-based feedback \cite{he2025tactile}. Despite these advances, the development of a calibrated sensing-to-action pipeline using bilateral antenna signals for reliable steering and recovery in an elongate multi-legged robot during traversal in narrow passages remains underexplored.

In this paper, we address this gap by developing a tactile antenna system that enables local, contact-based steering for elongate multi-legged robots in narrow passages (Fig. \ref{fig:demo}). Each antenna features a descending-stiffness design, allowing compliant probing while maintaining a stable signal under repeated contact. Through experimental calibration, we show that antenna deformation can be mapped to low-dimensional variables, including discrete bilateral contact states suitable for locomotion decisions. We further evaluate sensing performance across structured wall geometries and traversal speeds. By turning physical contact into signals that directly guide control, the proposed antenna provides a robust interface between environmental interaction and locomotion control, improving autonomy in elongate multi-legged robots without relying on computation-heavy perception or hardware-intensive sensing modalities.

\section{Hardware Design}
\subsection{Antenna morphology and function in \textit{S. subspinipes}}
Centipede antennae are built for repeated contact sensing during locomotion.
Using scanning electron microscopy (SEM), prior work characterized the external morphology of scolopendrid antennae and mapped the structure and distribution of antennal sensilla along the antennomeres \cite{ernst2013structure}.
A scolopendrid antenna is a slender, segmented chain composed of many short antennal segments connected by thin, movable joint membranes, which allows local bending while maintaining reach during contact.
The same SEM study revealed that the antenna is not uniform along its length: the diameter decreases toward the distal end and sensory structures are distributed non-uniformly as shown Fig. \ref{fig:niceshot}, with higher sensilla density in more distal segments and the highest density at the terminal segment \cite{ernst2013structure}. For our purposes, the main implication is simple. The distal region is specialized for probing contact, while the proximal region supports the structure and transmits deformation. This supports the design idea used in the robot, where antenna mechanics convert distributed contact into compact signals that guide locomotion decisions.

\begin{figure}
    \centering
    \includegraphics[width=0.95\linewidth]{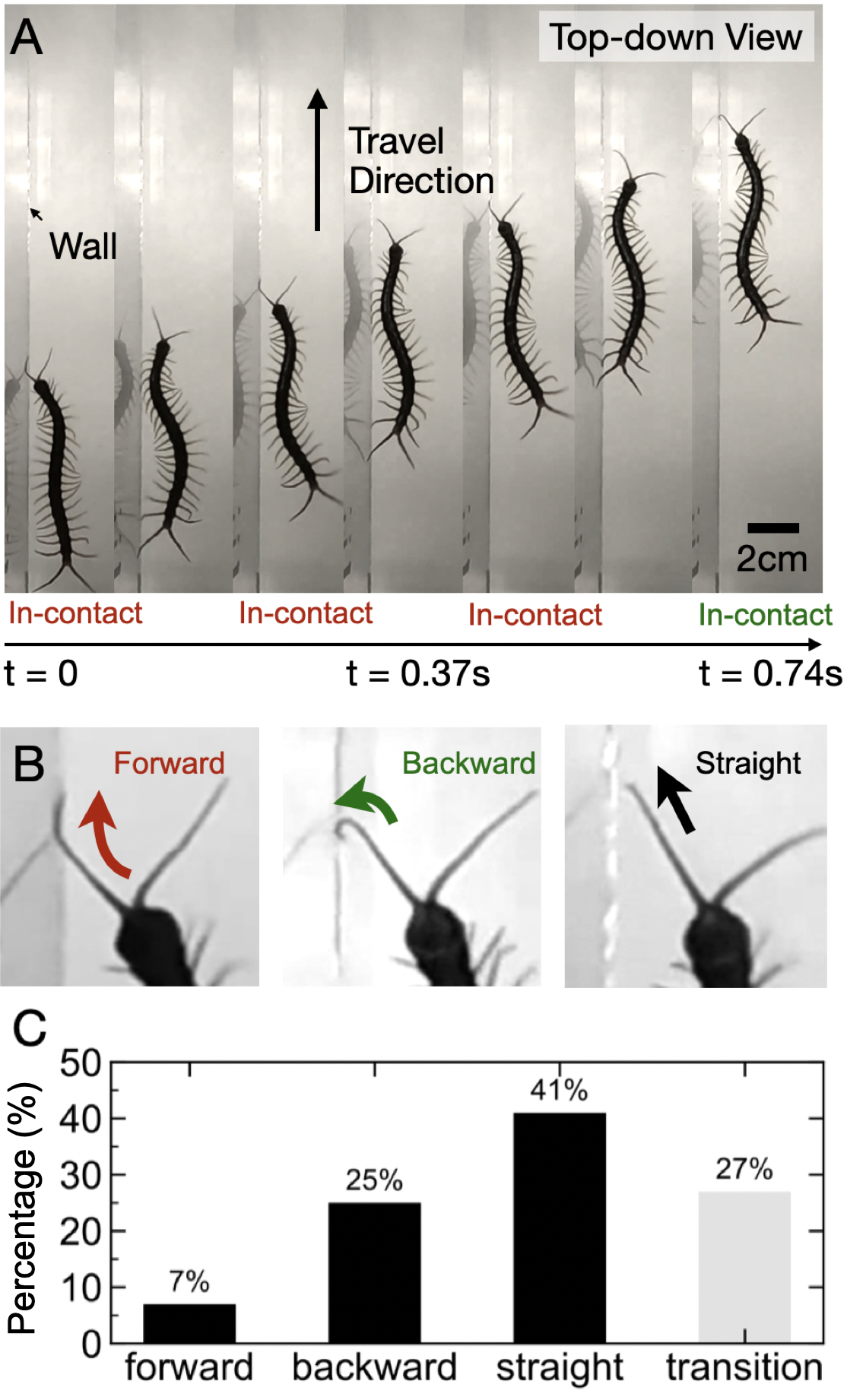}
    \caption{
    \textbf{Antenna contact and in-contact antenna morphologies during wall-following in \textit{S. subspinipes} (top-down view).}
    (A) Representative time sequence showing sustained antenna contact with a wall during forward travel (scale bar: 2\,cm).
    (B) Three stereotyped in-contact antenna morphologies, defined by the direction the antenna points relative to the body: \textit{forward pointing}, \textit{backward pointing}, and \textit{straight}.
    (C) Distribution of in-contact antenna morphologies across trials ($n=4$), including a \textit{transition} category for brief switching between morphologies during continuous contact.
    }
    \label{fig:animal_antenna}
\end{figure}

Motivated by this anatomy, prior work studied locomotion of \textit{S. polymorpha} in obstacle-rich lattice environments. \cite{pierce2026legged} Here, we focus on wall-following in \textit{S. subspinipes} and quantify antennal shape during sustained contact (Fig.~\ref{fig:animal_antenna}).
In these trials, antenna contact persists across locomotion cycles and is associated with repeatable locomotion pattern.
Fig.~\ref{fig:animal_antenna}B illustrates three stereotyped antenna morphologies while in contact, labeled by the direction the antenna points relative to the body: \textit{forward pointing}, \textit{backward pointing}, and \textit{straight}.
Fig.~\ref{fig:animal_antenna}C summarizes the observed distribution of these in-contact morphologies and a \textit{transition} category that captures brief switching between them during continuous contact.

\subsection{Descending stiffness in S. subspinipes antenna}
To quantify the mechanical gradient observed in the animal, we model the biological antenna as a sequence of $N_\mathrm{seg}=14$ discrete segments from the proximal base ($j=1$) to the distal tip ($j=14$), corresponding to the morphology in Fig.~\ref{fig:niceshot}B and C. 

\begin{figure}
    \centering
    \includegraphics[width=1\linewidth]{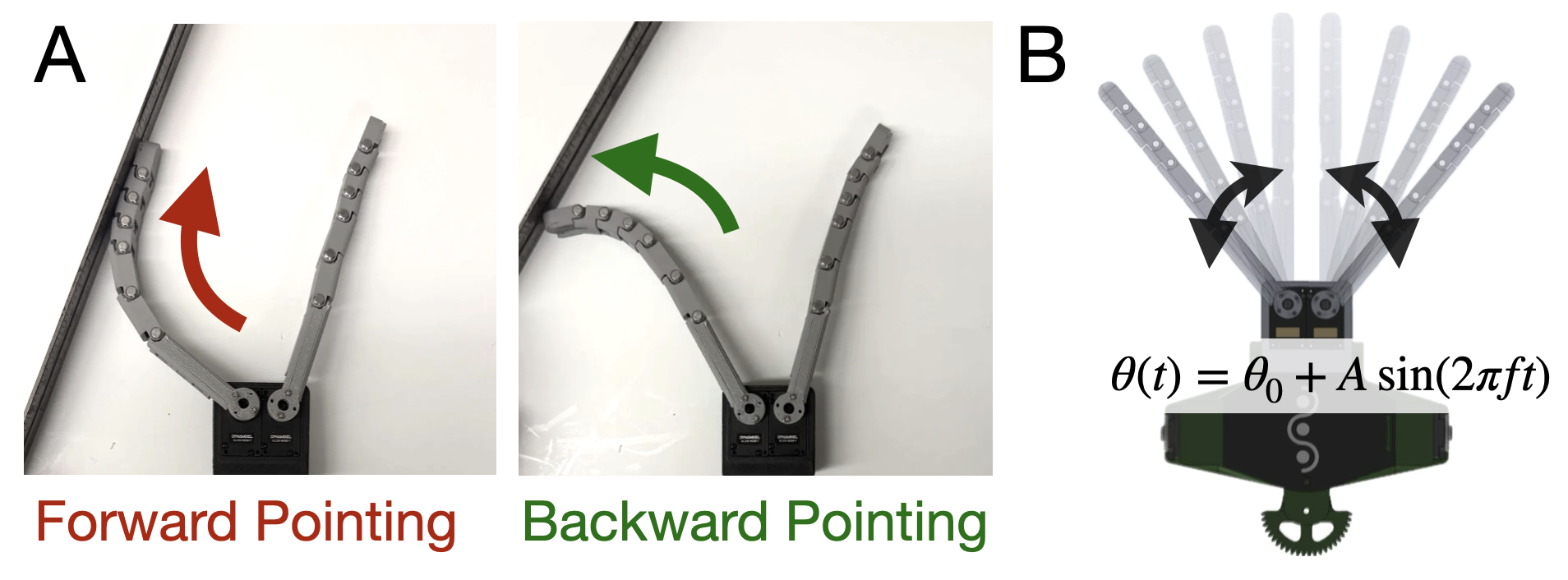}
    \caption{
    \textbf{Robotic antenna modes and actuation schematic.}
    (A) Representative in-contact antenna morphologies during wall interaction, labeled by the direction the antenna points relative to the body (\textit{forward pointing} and \textit{backward pointing}).
    (B) Planar antenna sweep used for calibration and testing, parameterized by a sinusoidal base angle $\theta_\mathrm{joint}(t)=\theta_0+A\sin(2\pi f t)$.
}
\label{fig:robot_antenna_modes}
\end{figure}
If each segments is approximated as a circular beam with roughly uniform material $E$ and similar segment length, then the local bending stiffness scales like $EI$, with the area moment of inertia given by:
\begin{equation}
    I = \frac{\pi d^4}{64}
\end{equation}
where $d$ is the local segment diameter. Because the area moment of inertia $I$ depends on the fourth power of the diameter, the local bending stiffness $K$ is strictly proportional to $d^4$. 

To evaluate the relative compliance profile along the structure, we define a normalized stiffness for the $j$-th segment relative to the rigid proximal base ($j=1$):
\begin{equation}
    \hat{K}_j = \left(\frac{d_j}{d_1}\right)^4
\end{equation}

As shown in Fig.~\ref{fig:niceshot}C, the diameter taper produces a descending stiffness profile along the antenna. The proximal segments remain relatively rigid to support the structure, while stiffness decreases toward the distal tip. We translate this structural principle directly into the robotic hardware.

\subsection{Antenna hardware}
\subsubsection{Sensing element and electronics}
Each antenna measures bending using a resistive flex sensor (ZD10-100) mounted along the antenna backbone.
The flex sensor changes resistance with curvature, producing a scalar signal that tracks antenna deflection.
We convert resistance to voltage using a voltage divider and sample the divider output using the ESP32 ADC.
The ESP32 streams time-stamped measurements for calibration and controller tuning, and supports wireless communication during experiments.

\subsubsection{Data acquisition and sampling}
The ESP32 samples the left and right antenna channels as raw ADC counts, $r_L(t)$ and $r_R(t)$, at a fixed sampling rate $f_s$.
Unless otherwise noted, we use $f_s \approx 100$~Hz (set by loop timing) and stream data over serial. To reduce sensor noise while preserving contact transients, we compute a short-time average over a window $T_{\mathrm{avg}}$ (Eq.~\ref{eq:avg}).
These averaged signals are the inputs to the calibration map and the discrete contact-state logic used in the controller.

\subsubsection{Mechanical integration and actuation}
The antenna is mounted on the anterior segment of SCUTTLE (Ground Control Robotics, Inc.) using a modular bracket that fixes the base pose relative to the body and allows rapid replacement after repeated tests. The flex sensor is mechanically protected by placing it close to the backbone and covering it with a thin protective layer to reduce abrasion during sliding contact.
For controlled characterization, we actuate the antenna base in the plane using a prescribed sweep motion (Fig.~\ref{fig:robot_antenna_modes}B) with two bus servos DYNAMIXEL XL330-M288-T, which provides repeatable deflection trajectories for calibration and experiments. The same hardware also produces the in-contact morphologies observed during wall interaction (Fig.~\ref{fig:robot_antenna_modes}A).
\subsubsection{Descending stiffness}

The antenna must survive repeated collisions and frictional sliding contact during traversal. To provide compliance while protecting the electronics and sensor, we implement a descending-stiffness antenna using torsional springs distributed along the antenna from base to tip.

Each spring can be modeled as a close-coiled helical torsion spring with a linear torque–angle relation:
\begin{equation}
    T(\Theta) = k_{\mathrm{torsion}}\,\Theta,
    \label{eq:torque_angle}
\end{equation}
where the torsional stiffness is approximated by:
\begin{equation}
    k_{\mathrm{torsion}} = \frac{E d_w^{4}}{10.8\,D\,N_c}.
    \label{eq:stiffness}
\end{equation}

Here, $E$ is Young's modulus, $d_w$ is the wire diameter, $D$ is the mean coil diameter, and $N_c$ is the number of active coils. The stiffness scales approximately as $k_{\mathrm{torsion}} \propto 1/N_c$, so fewer coils produce a stiffer response, and the constant $10.8$ arises from curved-beam bending theory.

In this work, we use six stainless steel torsion springs distributed along the antenna: one stiff spring at the base ($N_c=3$ coils), two medium springs in the middle ($N_c=6$ coils), and three compliant springs near the tip ($N_c=9$ coils). During operation, the antenna deflection remains below $30^\circ$, ensuring linear elastic behavior. The theoretical stiffness values for these distinct configurations are summarized in Table~\ref{tab:stiffness_values}.

\begin{table}[htbp]
    \centering
    \caption{Theoretical Torsional Stiffness by Active Coils}
    \label{tab:stiffness_values}
    \begin{tabular}{@{}ccc@{}}
        \toprule
        \textbf{Active Coils ($N_c$)} & \textbf{$k_{\mathrm{torsion}}$ (N$\cdot$mm/deg)} & \textbf{Position} \\ 
        \midrule
        3 &0.274 & Base (Stiffest) \\ 
        6 &  0.137 & Middle (Moderate) \\ 
        9 &  0.091 & Tip (Compliant) \\ 
        \bottomrule
    \end{tabular}
\end{table}

\begin{figure}[t]
    \centering
    \includegraphics[width=1.0\linewidth]{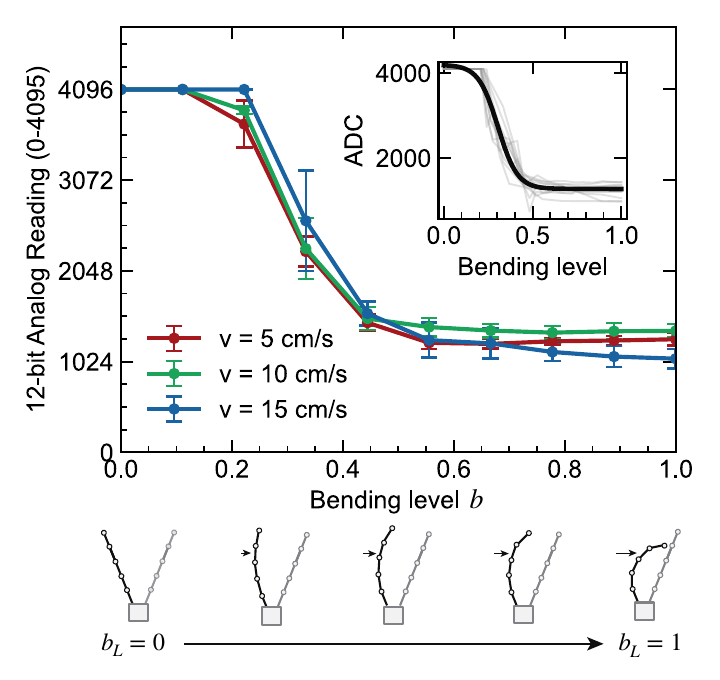}
    \caption{
    \textbf{Antenna calibration and mapping from ADC to normalized bending level.}
    Raw 12-bit ADC readings are recorded while the antenna is swept through controlled deflections at three traverse speeds ($v=5,10,15$\,cm/s).
    Markers show the mean and error bars indicate variability across trials.
    A global decreasing sigmoid fit (inset) defines the mapping from averaged ADC to a normalized bending level $b\in[0,1]$ used by the controller.
    Fit parameters for this dataset:  $k_s=18.3$, $x_0=0.305$.
    }
    \label{fig:calibration}
\end{figure}

\begin{figure*}[!t]
    \centering
    \includegraphics[width=1.0\linewidth]{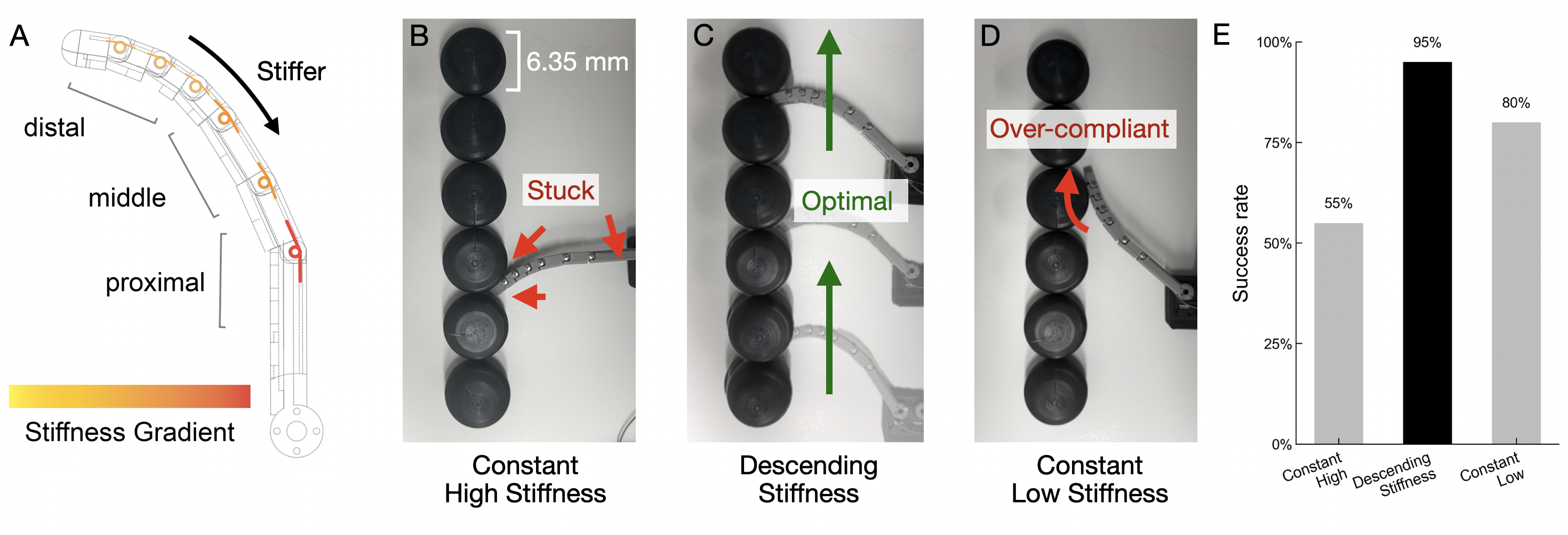}
    \caption{\textbf{Effect of antenna compliance on contact interaction.}
(A) Interchangeable torsion springs set three stiffness conditions (stiff: $N_c{=}3$, $k_{\mathrm{torsion}}{=}0.274$~N$\cdot$mm/deg; medium: $N_c{=}6$, $k_{\mathrm{torsion}}{=}0.137$~N$\cdot$mm/deg; compliant: $N_c{=}9$, $k_{\mathrm{torsion}}{=}0.091$~N$\cdot$mm/deg).
Too rigid (B) increases contact forces and causes jamming and too compliant (D) leads to collapse and missed contact; the smooth descending stiffness profile (C) is optimal with high temporal resolution reading without getting stuck.
(E) Bar chart summarizes traversal success rate for the three stiffness profile.}
    \label{fig:stiffness}
\end{figure*}

\section{Antenna performance}
\subsection{Calibration and signal processing}
We calibrate each antenna channel using a controlled indentation experiment with a robot arm.
The antenna is passive during calibration (no base actuation): a rigid end-effector pushes the antenna laterally to impose repeatable deflections while we record the raw 12-bit ADC output from the flex sensor.
We repeat the procedure across multiple pushing speeds to capture rate-dependent effects such as frictional sliding and viscoelastic response.
Across experiments with varied tested speeds, the response is not time independent but strongly deformation dependent, allowing us to define a single ground-truth calibration curve that maps sensor output to normalized bending level (Fig.~\ref{fig:calibration}).

During operation, the controller uses the same sensing front-end.
We compute short-time averaged antenna signals over a window $T_{\mathrm{avg}}$,
\begin{equation}
\bar{r}_i(t)=\frac{1}{T_{\mathrm{avg}}}\int_{t-T_{\mathrm{avg}}}^{t} r_i(\tau)\,d\tau,\quad i\in\{L,R\}.
\label{eq:avg}
\end{equation}
We then map $\bar{r}_i(t)$ to a normalized bending/contact level $b_i(t)\in[0,1]$ using a sigmoid calibration,
\begin{equation}
b_i(t)=\sigmoid{k_s}{\bar{r}_i(t)}{r_0}.
\label{eq:sigmoid}
\end{equation}
Here $k_s$ controls the slope and $r_0$ is the midpoint.
For logic-based decisions, we define a contact threshold $\theta$ and a binary contact state,
\begin{equation}
c_i(t) = \big(b_i(t)\ge \theta\big),\quad i\in\{L,R\}.
\label{eq:contact}
\end{equation}
These variables $(b_L,b_R,c_L,c_R)$ are used for controller decisions in narrow-space traversal.

\subsection{Robustness Analysis}

We quantify how antenna stiffness profile affects contact robustness using a repeatable boulder-wall traversal test (Fig.~\ref{fig:stiffness}).
The antenna base angle is fixed throughout the test, and the antenna is passive without actuation. A robot arm drives the antenna along the wall at a constant speed of 8\,mm/s to standardize contact conditions across trials. The wall is constructed from 6.35\,mm diameter spherical boulders arranged in a vertical line, producing repeated discrete contacts along the antenna (Fig.~\ref{fig:stiffness}B-D).

We compare three stiffness profiles: constant high stiffness, constant low stiffness, and a descending stiffness design (Fig.~\ref{fig:stiffness}A). Each condition is tested for $n=20$ trials, and a trial is considered successful if the antenna traverses the wall without jamming or stuck. As shown in Fig.~\ref{fig:stiffness}E, constant high stiffness is jamming at boulder contacts, while constant low stiffness is over-compliant and tends to slip, reducing the ability to maintain a consistent contact for sensor reading.  

\section{Robotic Experiments}

\subsection{Hardware platform}
Experiments are performed on SCUTTLE (Ground Control Robotics, Inc.), a commercialized multi-legged elongate robot platform designed for low-profile traversal. This robot has a 6-link body with five actuated inter-link yaw joints and 12 legs (six contralateral leg pairs). Each leg is O-leg shape actuated by a dedicated servo (XC430-W240-T), and the body joints are actuated by the same bus servos. 

The platform supports both planar (yaw) body undulation and additional vertical (pitch) body degrees of freedom. In this paper, we use only planar lateral undulation of “serpenoid” wave\cite{hirose1993biologically} to generate gait and steering maneuvers in 2D planar settings.

We prescribe a traveling-wave body undulation for the $i$-th yaw joint angle $\alpha_i$ at time $t$ as
\begin{equation}
\alpha_i(t)=A_{\mathrm{body}}\,\sin\!\left(2\pi\,\xi\,\frac{i}{N_{i}}-2\pi\,\omega\,t\right)+\varphi,
\label{eq:serpenoid}
\end{equation}
where $A_{\mathrm{body}}$ is the amplitude, $\xi$ is the spatial frequency (waves along the body), $\omega$ is the temporal frequency, $\varphi$ is an angle offset, $i$ is the joint index, and $N_i$ is the total number of yaw joints.

The robot is powered by a DC power supply at 11.1\,V and receives control commands from a PC via a U2D2 serial adapter (ROBOTIS). Power and TTL communication share the same tether, and the servo wiring is routed through the body modules so that all the motors are connected along a single daisy-chained bus. Antenna servos are commanded over the shared TTL bus whereas, for the XL330 actuators, we use a 5\,V step-down (buck) module to provide the required supply voltage from the 11.1\,V input.

\subsection{Narrow and complex tunnel traversal}
\begin{figure}
    \centering
    \includegraphics[width=0.86\linewidth]{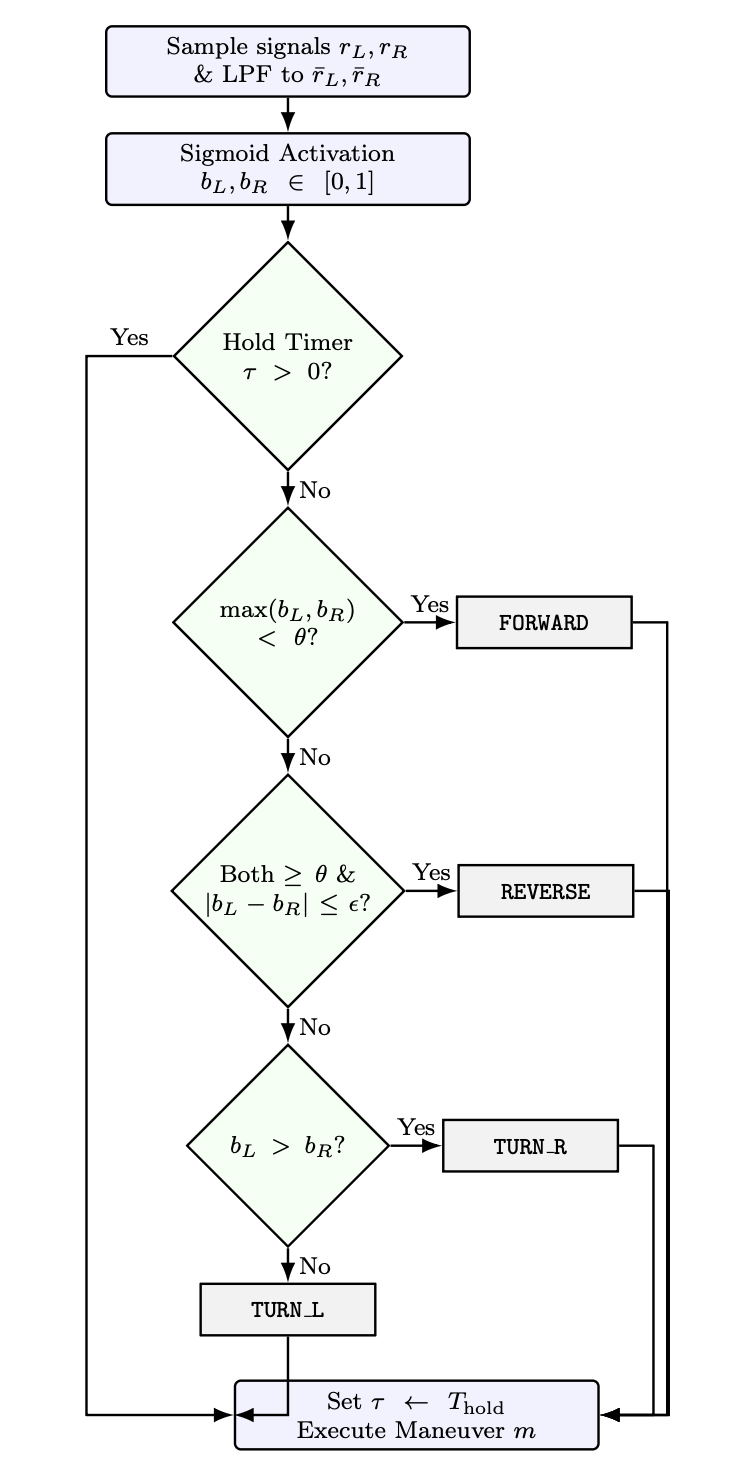}
    \caption{\textbf{Flowchart of the discrete antenna-based controller for confined traversal.} Bilateral antenna signals are mapped through a sigmoid to contact variables, which drive the decision process with a time window holding. The controller selects among forward, turn, and reverse maneuvers based on contact presence, symmetry, and dominance to provide close-loop navigation in confined space.}
    \label{fig:flowchart}
\end{figure}

The robot uses a discrete controller (Fig. \ref{fig:flowchart}) that maps bilateral antenna contact into one of four maneuvers,
$m\in\{\texttt{FORWARD},\texttt{TURN\_L},\texttt{TURN\_R},\texttt{REVERSE}\}$, and holds the selected maneuver for a fixed duration $T_{\mathrm{hold}}$.
Each maneuver is implemented by switching the open-loop serpenoid body-wave template in Eq.~\ref{eq:serpenoid}:
\texttt{FORWARD} uses the head-to-tail traveling wave, \texttt{TURN\_L}/\texttt{TURN\_R} use the same wave with an offset to create a left/right yaw, and \texttt{REVERSE} uses a reverse template to back out when the robot is stuck.
Decisions use the ground truth antenna signals from calibration: (Eqs.~\ref{eq:avg}-\ref{eq:contact}):
if neither antenna indicates contact, the controller commands \texttt{FORWARD};
if both sides indicate strong and similar contact, it commands \texttt{REVERSE};
otherwise it turns away from the contacting side.

\begin{figure}
    \centering
    \includegraphics[width=1\linewidth]{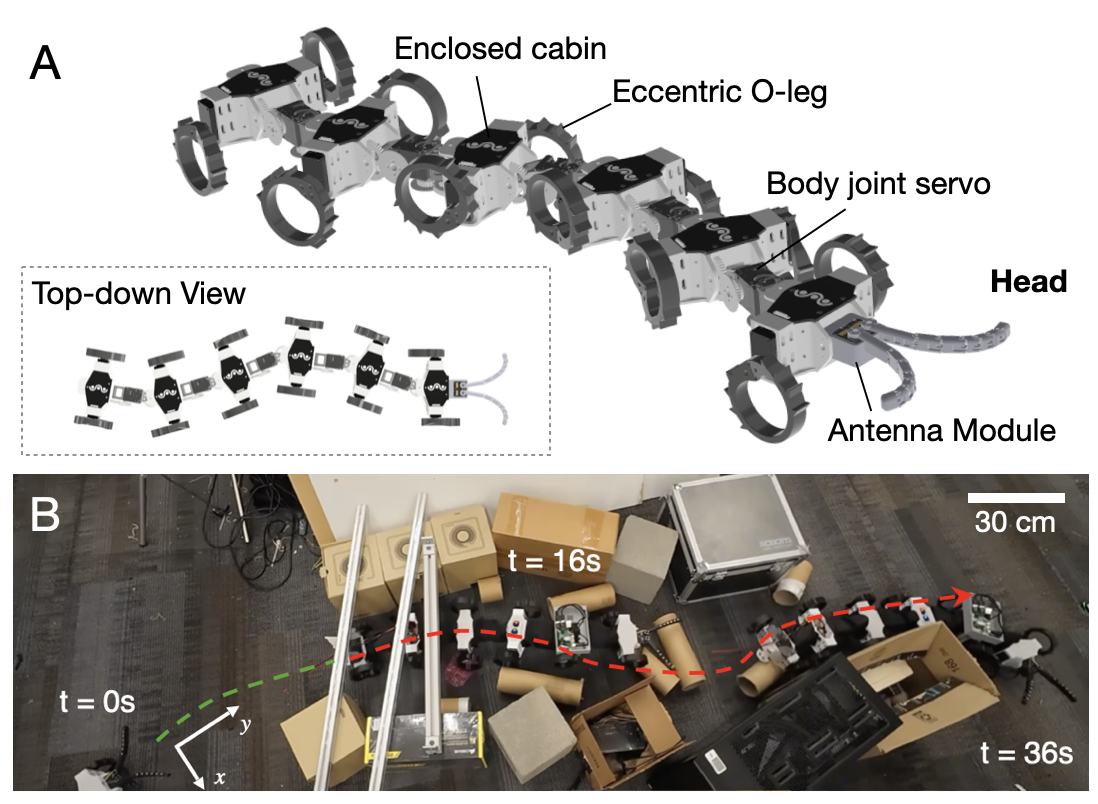}
    \caption{\textbf{Experiment on robotic platform SCUTTLE.} (A) Modular multi-legged robot with enclosed cabin, eccentric O-legs, body joint servos, and a bilateral antenna module (top-down view inset showing traveling-wave gait).
    (B) Overhead view of cluttered traversal showing representative trajectory with close-loop control.}
    \label{fig:demo}
\end{figure}

Fig.~\ref{fig:demo}B shows the experimental setup and a representative run through a narrow tunnel with a turn, where contact is frequent. The tunnel is filled with lightweight, shifting paper cylinders (7~cm radius), and the widest local section is 40~cm across—about twice the robot’s width. In the tunnel, one-sided contact triggers a turning template that steers the head away from the wall and helps the body re-center, while symmetric contact triggers a short reverse to reduce contact and regain clearance.

The feedback controller achieves a 100\% success rate across all 10 trials, with an average speed of 0.12 m/s. Under open-loop control, the success rate drops to 60\% (10 trials total), and the successful runs average 0.04 m/s. Across repeated trials, the closed-loop strategy is robust and maintains a centered trajectory through the turn, whereas the same gait in open loop frequently becomes stuck.

Across all trials, we observed no antenna failures, even under repeated sliding contact and collisions.

\section{Conclusion and Discussion}

In this paper, we presented a tactile antenna system for SCUTTLE that implements a robophysical model of centipede antenna mechanics. Motivated by the morphology of centipedes, the core contribution of this work is the antenna's descending stiffness profile. By distributing torsion springs from a stiff base to a compliant tip, the antenna reliably negotiates sliding contacts and collisions. It avoids the jamming typical of rigid sensors and the signal loss typical of compliant ones. We mapped the raw flex sensor data to a normalized bending scale and implemented a discrete feedback controller. Our experiments demonstrated that this easy-to-implement and low-cost approach allows the robot to reliably steer through narrow, cluttered environment and recover from wedged states where open-loop locomotion fails.

These results highlight the value of mechanical intelligence~\cite{wang2023mechanical,kortman2025perspectives,khaheshi2022mechanical} in high degree-of-freedom elongate robots. Mechanical intelligence has already proven powerful in open-loop locomotion, where morphology and passive dynamics enhance robustness with minimal control. Here, we extend that idea to contact-based navigation. By introducing mechanically intelligent antenna compliance, the mechanics shape body–environment interactions into stable feedback signals. This reduces the burden on the computational cost and demonstrates that robust confined-space traversal does not strictly require complex global mapping or real-time vision. 

For future work, we aim to expand our control framework. While the current discrete bang-bang controller is empirically robust, it is limited to a small set of predefined maneuvers. We will integrate these tactile states with geometric mechanics to optimize gaits in real time and we expect the robot can dynamically change its gait to smoothly traverse through various irregular environments autonomously. 

Finally, biological centipedes not only perform rapid planar wall-following but also seamlessly transition over obstacles. Inspired by this, we will expand our testing into 3D environments. By adding bilateral antenna signals with vertical joint, the robot will be able to detect steps or missing, automatically triggering vertical pitch actuation to transition from 2D steering to spatial obstacle climbing. 
\section{Acknowledgement}
The authors would like to thank the Institute for Robotics and Intelligent Machines for use of the College of Computing basement as a testing space. The authors also thank Ground Control Robotics LLC for use of the robotic platform and technical support, and Daniel Soto for his support and assistance. This work was also supported by the NSF STTR Phase I grant (2335553) and Jim Pope Fellowship.

\bibliographystyle{IEEEtran}
\bibliography{main}

@article{chong2023multilegged,
  title={Multilegged matter transport: A framework for locomotion on noisy landscapes},
  author={Chong, Baxi and He, Juntao and Soto, Daniel and Wang, Tianyu and Irvine, Daniel and Blekherman, Grigoriy and Goldman, Daniel I},
  journal={Science},
  volume={380},
  number={6644},
  pages={509--515},
  year={2023},
  publisher={American Association for the Advancement of Science}
}

@article{aoi2022advanced,
  title={Advanced turning maneuver of a many-legged robot using pitchfork bifurcation},
  author={Aoi, Shinya and Tomatsu, Ryoe and Yabuuchi, Yuki and Morozumi, Daiki and Okamoto, Kota and Fujiki, Soichiro and Senda, Kei and Tsuchiya, Kazuo},
  journal={IEEE Transactions on Robotics},
  volume={38},
  number={5},
  pages={3015--3026},
  year={2022},
  publisher={IEEE}
}

@article{ozkan2021self,
  title={Self-reconfigurable multilegged robot swarms collectively accomplish challenging terradynamic tasks},
  author={Ozkan-Aydin, Yasemin and Goldman, Daniel I},
  journal={Science Robotics},
  volume={6},
  number={56},
  pages={eabf1628},
  year={2021},
  publisher={American Association for the Advancement of Science}
}

@article{pierce2026legged,
  title={Legged Locomotion in Lattices: Centipede Traversal of Obstacle-Rich Environments},
  author={Pierce, Christopher J and Soto, Daniel and Erickson, Eva and Diaz, Kelimar and Iaschi, Massimiliano and Lay, Anna and Goldman, Daniel I},
  journal={Annals of the New York Academy of Sciences},
  volume={1555},
  number={1},
  pages={e70187},
  year={2026},
  publisher={Wiley Online Library}
}

@article{chong2025omega,
  title={The Omega Turn: A General Turning Template for Elongate Robots},
  author={Chong, Baxi and Wang, Tianyu and Diaz, Kelimar and Pierce, Christopher J and Erickson, Eva and Whitman, Julian and Deng, Yuelin and Flores, Esteban and Fu, Ruijie and He, Juntao and others},
  journal={arXiv preprint arXiv:2510.12970},
  year={2025}
}

@article{he2025probabilistic,
  title={Probabilistic approach to feedback control enhances multi-legged locomotion on rugged landscapes},
  author={He, Juntao and Chong, Baxi and Lin, Jianfeng and Xu, Zhaochen and Bagheri, Hosain and Flores, Esteban and Goldman, Daniel I},
  journal={IEEE Transactions on Robotics},
  year={2025},
  publisher={IEEE}
}

@article{cowan2006task,
  title={Task-level control of rapid wall following in the American cockroach},
  author={Cowan, Noah J and Lee, Jusuk and Full, Robert J},
  journal={Journal of Experimental Biology},
  volume={209},
  number={9},
  pages={1617--1629},
  year={2006},
  publisher={Company of Biologists}
}

@article{full1999templates,
  title={Templates and anchors: neuromechanical hypotheses of legged locomotion on land},
  author={Full, Robert J and Koditschek, Daniel E},
  journal={Journal of experimental biology},
  volume={202},
  number={23},
  pages={3325--3332},
  year={1999},
  publisher={The Company of Biologists Ltd}
}

@article{he2025tactile,
  title={Tactile sensing enables vertical obstacle negotiation for elongate many-legged robots},
  author={He, Juntao and Chong, Baxi and Iaschi, Massimiliano and Nienhusser, Vincent R and Ha, Sehoon and Goldman, Daniel I},
  journal={arXiv preprint arXiv:2504.08615},
  year={2025}
}

@article{lee2008templates,
  title={Templates and anchors for antenna-based wall following in cockroaches and robots},
  author={Lee, Jusuk and Sponberg, Simon N and Loh, Owen Y and Lamperski, Andrew G and Full, Robert J and Cowan, Noah J},
  journal={IEEE Transactions on Robotics},
  volume={24},
  number={1},
  pages={130--143},
  year={2008},
  publisher={IEEE}
}

@article{mongeau2014mechanical,
  title={Mechanical processing via passive dynamic properties of the cockroach antenna can facilitate control during rapid running},
  author={Mongeau, Jean-Michel and Demir, Alican and Dallmann, Chris J and Jayaram, Kaushik and Cowan, Noah J and Full, Robert J},
  journal={Journal of Experimental Biology},
  volume={217},
  number={18},
  pages={3333--3345},
  year={2014},
  publisher={Company of Biologists}
}

@article{mongeau2013locomotion,
  title={Locomotion-and mechanics-mediated tactile sensing: antenna reconfiguration simplifies control during high-speed navigation in cockroaches},
  author={Mongeau, Jean-Michel and Demir, Alican and Lee, Jusuk and Cowan, Noah J and Full, Robert J},
  journal={Journal of Experimental Biology},
  volume={216},
  number={24},
  pages={4530--4541},
  year={2013},
  publisher={Company of Biologists}
}

@article{mcdonnell2025design,
  title={Design of a bioinspired robophysical antenna for insect-scale tactile perception and navigation},
  author={McDonnell, Parker and Meng, Lingsheng and Hariprasad, Hari Krishna and Hedrick, Alexander and Miscles, Eduardo and Gilinsky, Samuel and Mongeau, Jean-Michel and Jayaram, Kaushik},
  journal={arXiv preprint arXiv:2507.23719},
  year={2025}
}

@article{mongeau2015sensory,
  title={Sensory processing within cockroach antenna enables rapid implementation of feedback control for high-speed running maneuvers},
  author={Mongeau, Jean-Michel and Sponberg, Simon N and Miller, John P and Full, Robert J},
  journal={The Journal of Experimental Biology},
  volume={218},
  number={15},
  pages={2344--2354},
  year={2015},
  publisher={The Company of Biologists}
}

@article{camhi1999high,
  title={High-frequency steering maneuvers mediated by tactile cues: antennal wall-following in the cockroach},
  author={Camhi, JM and Johnson, EN},
  journal={Journal of Experimental Biology},
  volume={202},
  number={5},
  pages={631--643},
  year={1999},
  publisher={The Company of Biologists Ltd}
}

@article{okada2000role,
  title={The role of antennal hair plates in object-guided tactile orientation of the cockroach (Periplaneta americana)},
  author={Okada, J and Toh, Y},
  journal={Journal of Comparative Physiology A},
  volume={186},
  number={9},
  pages={849--857},
  year={2000},
  publisher={Springer}
}

@article{loudon2014cricket,
  title={Cricket antennae shorten when bending (Acheta domesticus L.)},
  author={Loudon, Catherine and Bustamante Jr, Jorge and Kellogg, Derek W},
  journal={Frontiers in Physiology},
  volume={5},
  pages={242},
  year={2014},
  publisher={Frontiers Media SA}
}

@inproceedings{lamperski2005dynamical,
  title={Dynamical wall following for a wheeled robot using a passive tactile sensor},
  author={Lamperski, Andrew G and Loh, Owen Y and Kutscher, Brett L and Cowan, Noah J},
  booktitle={Proceedings of the 2005 IEEE International Conference on Robotics and Automation},
  pages={3838--3843},
  year={2005},
  organization={IEEE}
}

@inproceedings{demir2010tunable,
  title={A tunable physical model of arthropod antennae},
  author={Demir, Alican and Samson, Edward W and Cowan, Noah J},
  booktitle={2010 IEEE International Conference on Robotics and Automation},
  pages={3793--3798},
  year={2010},
  organization={IEEE}
}

@article{pearson2011biomimetic,
  title={Biomimetic vibrissal sensing for robots},
  author={Pearson, Martin J and Mitchinson, Ben and Sullivan, J Charles and Pipe, Anthony G and Prescott, Tony J},
  journal={Philosophical Transactions of the Royal Society B: Biological Sciences},
  volume={366},
  number={1581},
  pages={3085--3096},
  year={2011},
  publisher={The Royal Society}
}

@article{solomon2006robotic,
  title={Robotic whiskers used to sense features},
  author={Solomon, Joseph H and Hartmann, Mitra J},
  journal={Nature},
  volume={443},
  number={7111},
  pages={525--525},
  year={2006},
  publisher={Nature Publishing Group UK London}
}

@article{evans2013effect,
  title={The effect of whisker movement on radial distance estimation: a case study in comparative robotics},
  author={Evans, Mathew H and Fox, Charles W and Lepora, Nathan F and Pearson, Martin J and Sullivan, J Charles and Prescott, Tony J},
  journal={Frontiers in neurorobotics},
  volume={6},
  pages={12},
  year={2013},
  publisher={Frontiers Media SA}
}

@article{kent2021whisksight,
  title={Whisksight: A reconfigurable, vision-based, optical whisker sensing array for simultaneous contact, airflow, and inertia stimulus detection},
  author={Kent, Teresa A and Kim, Suhan and Kornilowicz, Gabriel and Yuan, Wenzhen and Hartmann, Mitra JZ and Bergbreiter, Sarah},
  journal={IEEE Robotics and Automation Letters},
  volume={6},
  number={2},
  pages={3357--3364},
  year={2021},
  publisher={IEEE}
}

@article{zhu2023self,
  title={Self-powered bionic antenna based on triboelectric nanogenerator for micro-robotic tactile sensing},
  author={Zhu, Dekuan and Lu, Jiangfeng and Zheng, Mingjie and Wang, Dongkai and Wang, Jianyu and Liu, Yixin and Wang, Xiaohao and Zhang, Min},
  journal={Nano Energy},
  volume={114},
  pages={108644},
  year={2023},
  publisher={Elsevier}
}

@inproceedings{shahmiri2020sharc,
  title={Sharc: A geometric technique for multi-bend/shape sensing},
  author={Shahmiri, Fereshteh and Dietz, Paul H},
  booktitle={Proceedings of the 2020 CHI Conference on Human Factors in Computing Systems},
  pages={1--12},
  year={2020}
}

@article{yu2022bioinspired,
  title={Bioinspired, multifunctional, active whisker sensors for tactile sensing of mobile robots},
  author={Yu, Zhiqiang and Guo, Yue and Su, Jiaji and Huang, Qiang and Fukuda, Toshio and Cao, Changyong and Shi, Qing},
  journal={IEEE Robotics and Automation Letters},
  volume={7},
  number={4},
  pages={9565--9572},
  year={2022},
  publisher={IEEE}
}

@article{ernst2013structure,
  title={Structure and distribution of antennal sensilla in the centipede Scolopendra oraniensis (Lucas, 1846)(Chilopoda, Scolopendromorpha)},
  author={Ernst, Alfred and Hilken, Gero and Rosenberg, J{\"o}rg and Voigtl{\"a}nder, Karin and Sombke, Andy},
  journal={Zoologischer Anzeiger-A Journal of Comparative Zoology},
  volume={252},
  number={2},
  pages={217--225},
  year={2013},
  publisher={Elsevier}
}

@book{hirose1993biologically,
  title={Biologically inspired robots: serpentile locomotors and manipulators},
  author={Hirose, Shigeo},
  year={1993},
  publisher={Oxford University Press, Inc.}
}

@article{chong2023self,
  title={Self-propulsion via slipping: Frictional swimming in multilegged locomotors},
  author={Chong, Baxi and He, Juntao and Li, Shengkai and Erickson, Eva and Diaz, Kelimar and Wang, Tianyu and Soto, Daniel and Goldman, Daniel I},
  journal={Proceedings of the National Academy of Sciences},
  volume={120},
  number={11},
  pages={e2213698120},
  year={2023},
  publisher={National Academy of Sciences}
}

@article{chong2022general,
  title={A general locomotion control framework for multi-legged locomotors},
  author={Chong, Baxi and O Aydin, Yasemin and Rieser, Jennifer M and Sartoretti, Guillaume and Wang, Tianyu and Whitman, Julian and Kaba, Abdul and Aydin, Enes and McFarland, Ciera and Diaz Cruz, Kelimar and others},
  journal={Bioinspiration \& Biomimetics},
  volume={17},
  number={4},
  pages={046015},
  year={2022},
  publisher={IOP Publishing}
}

@article{yu2024whisker,
  title={Whisker sensor for robot environments perception: a review},
  author={Yu, Zhenhua and Childs, Peter RN and Ge, Yukun and Nanayakkara, Thrishantha},
  journal={IEEE Sensors Journal},
  volume={24},
  number={18},
  pages={28504--28521},
  year={2024},
  publisher={IEEE}
}

@article{pearson2007whiskerbot,
  title={Whiskerbot: a robotic active touch system modeled on the rat whisker sensory system},
  author={Pearson, Martin J and Pipe, Anthony G and Melhuish, Chris and Mitchinson, Ben and Prescott, Tony J},
  journal={Adaptive Behavior},
  volume={15},
  number={3},
  pages={223--240},
  year={2007},
  publisher={Sage Publications Sage UK: London, England}
}

@article{wang2023mechanical,
  title={Mechanical intelligence simplifies control in terrestrial limbless locomotion},
  author={Wang, Tianyu and Pierce, Christopher and Kojouharov, Velin and Chong, Baxi and Diaz, Kelimar and Lu, Hang and Goldman, Daniel I},
  journal={Science Robotics},
  volume={8},
  number={85},
  pages={eadi2243},
  year={2023},
  publisher={American Association for the Advancement of Science}
}

@article{khaheshi2022mechanical,
  title={Mechanical intelligence (MI): a bioinspired concept for transforming engineering design},
  author={Khaheshi, Ali and Rajabi, Hamed},
  journal={Advanced Science},
  volume={9},
  number={32},
  pages={2203783},
  year={2022},
  publisher={Wiley Online Library}
}

@article{kortman2025perspectives,
  title={Perspectives on intelligence in soft robotics},
  author={Kortman, Vera Gesina and Mazzolai, Barbara and Sakes, Aime{\'e} and Jovanova, Jovana},
  journal={Advanced Intelligent Systems},
  volume={7},
  number={1},
  pages={2400294},
  year={2025},
  publisher={Wiley Online Library}
}

\end{document}